\documentclass[twocolumn]{bmcart}

\usepackage{amsthm,amsmath}
\RequirePackage{hyperref}
\usepackage[colorlinks,linkcolor=blue]{hyperref}
\usepackage[utf8]{inputenc} 
\usepackage{amsfonts}
\usepackage{graphicx}
\usepackage{diagbox}
\usepackage{subfigure}

\startlocaldefs
\endlocaldefs

\begin{document}

\begin{frontmatter}

\begin{fmbox}
\dochead{Research}


\title{Deep Superpixel-based Network for Blind Image Quality Assessment}


\author[
  addressref={aff1},                   
  corref={aff1},                       
 email={ygy@whu.edu.cn}   
]{\inits{GY}\fnm{Guangyi} \snm{Yang}}
\author[
  addressref={aff1},
 email={johnyoung@whu.edu.cn}
]{\inits{ZY}\fnm{Yang} \snm{Zhan}}
\author[
addressref={aff1},
 email={john.RS.Smith@cambridge.co.uk}
]{\inits{YX}\fnm{Yuxuan} \snm{Wang}}


\address[id=aff1]{
	\orgname{School of Electronic Information, Wuhan University}, 
	\postcode{430072}                          
	\city{Wuhan},                              
	\cny{China}                                
}





\begin{abstractbox}

\begin{abstract} 
The goal in a blind image quality assessment (BIQA) model is to simulate the process of evaluating images by human eyes and accurately assess the quality of the image. Although many approaches effectively identify degradation, they do not fully consider the semantic content in images resulting in distortion. In order to fill this gap, we propose a deep adaptive superpixel-based network, namely DSN-IQA, to assess the quality of image based on multi-scale and superpixel segmentation. The DSN-IQA can adaptively accept arbitrary scale images as input images, making the assessment process similar to human perception. The network uses two models to extract multi-scale semantic features and generate a superpixel adjacency map. These two elements are united together via feature fusion to accurately predict image quality. Experimental results on different benchmark databases demonstrate that our algorithm is highly competitive with other approaches when assessing challenging authentic image databases. Also, due to adaptive deep superpixel-based network, our model accurately assesses images with complicated distortion, much like the human eye.
\end{abstract}




\end{abstractbox}
\end{fmbox}

\end{frontmatter}



\section{Introduction} \label{sect:intro}
The unprecedented development of communication technologies has underscored the role of images as the main carrier of visual information~\cite{yan20203d}. In many situations, the quality of an image is correlated to the coherence of the content since distortions have a significant negative impact on the readability of an image. Almost every stage in image acquisition, transmission, and storage could cause different degrees of distortion~\cite{Yang2013subjective}. Consequently, image quality assessment (IQA) is necessary for monitoring the quality of images and thus assuring the reliability of image processing systems. As a result, research on IQA has received wide attention.

IQA methods can be divided into two types: subjective assessment and objective assessment, depending whether they need human eyes for classification~\cite{wangmodern}. Subjective assessment evaluates images using human eyes~\cite{zhang2009image} and based on intuitive visual experience is the standard. It is the most accurate metric because the perceptual image quality is in the final instance, judged by the human visual system (HVS). However, it is time-consuming, expensive and the practical tasks are too laborious for routine implementation. So, objective assessment is more practical and widely used because a machine can automatically predict the quality of an image using mathematical models. Objective assessment is usually divided in to three categories on account of the presence or absence of a reference image: full-reference IQA (FR-IQA), reduced-reference IQA (RR-IQA) and no-reference or blind IQA (BIQA)~\cite{wangmodern}. In practical application processes, a reference image is often not given, which impedes the application scope of FR-IQA and RR-IQA. BIQA therefore, has attracted growing interest among researchers~\cite{wang2011reduced}. 

\begin{figure*}[h!]
	\centering
	\subfigure[Original image]{
		\begin{minipage}[t]{0.23\textwidth}
			\centering
			\includegraphics[width=0.95\textwidth]{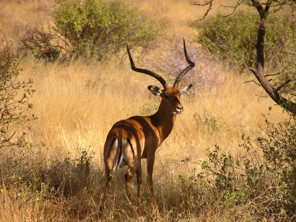}
		\end{minipage}%
	}%
	\subfigure[Crop]{
		\begin{minipage}[t]{0.23\textwidth}
			\centering
			\includegraphics[width=0.95\textwidth]{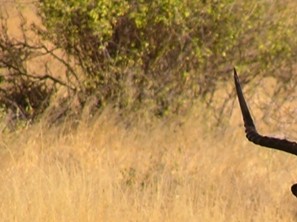}
		\end{minipage}%
	}
	\subfigure[Rotate]{
	\begin{minipage}[t]{0.23\textwidth}
		\centering
		\includegraphics[width=0.95\textwidth]{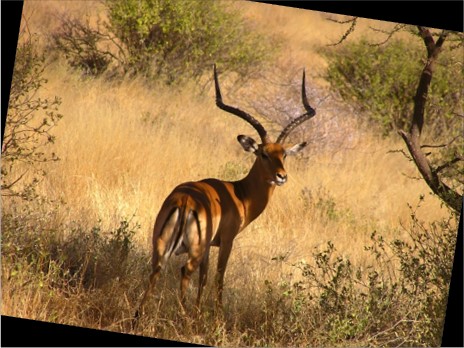}
	\end{minipage}%
	}
	\subfigure[Rescale]{
	\begin{minipage}[t]{0.165\textwidth}
		\centering
		\includegraphics[width=0.99\textwidth]{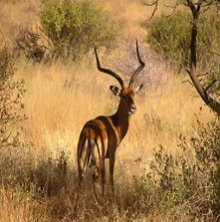}
	\end{minipage}%
	}

	\subfigure[Original image]{
	\begin{minipage}[t]{0.23\textwidth}
		\centering
		\includegraphics[width=0.95\textwidth]{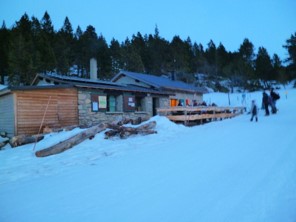}
	\end{minipage}%
	}
	\subfigure[Crop]{
	\begin{minipage}[t]{0.23\textwidth}
		\centering
		\includegraphics[width=0.95\textwidth]{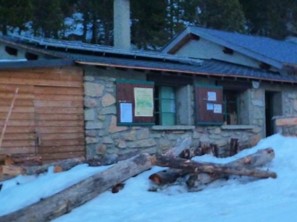}
	\end{minipage}%
	}
	\subfigure[Crop]{
	\begin{minipage}[t]{0.23\textwidth}
		\centering
		\includegraphics[width=0.95\textwidth]{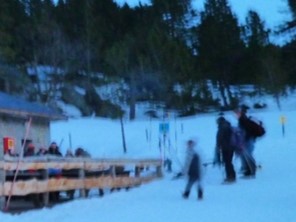}
	\end{minipage}%
	}
	\subfigure[Rescale]{
	\begin{minipage}[t]{0.165\textwidth}
		\centering
		\includegraphics[width=0.99\textwidth]{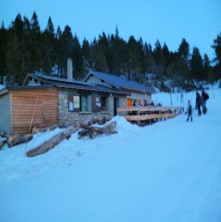}
	\end{minipage}%
	}
	\caption{Pre-processing ways change the quality of image, including cropping, rotation, rescaling. (a), (e)Original images. (b), (f)-(g)Cropped images. Levels of the distortion and contents visually change due to the cropping. (c)Rotated image. The unreal black is introduced to pad the edges. (d), (h)Rescaled images, which contain conspicuous deformation.}
	\label{figure1}
\end{figure*}

Feature extraction algorithms make the BIQA model more widely applicable; however, the current algorithms explicitly designed for the model have weakness. Some BIQA models adopt low-level features and employ machine learning to assess quality: using a learning-based regression model trained by a set of features extracted from training images whose mean opinion scores (MOS) or Different MOS (DMOS) already gained via subjective experiments. This regression model is then used to predict the ground truth MOS. The BIQA models apply the principles of natural scene statistics (NSS) and can successfully represent the overall quality of the image. However, they are not effective for evaluating local distortions in an image. To resolve this issue feature extractors based on deep convolutional neural networks (CNN)~\cite{kang2014convolutional, ma2017end} have been proposed and are widely used among researchers. They automatically capture deep features to represent degradation and they have been widely used in BIQA tasks. One of the major problems with CNN-based image quality assessment however, is that the attention of our eyes to the image is not evenly distributed across different regions. Ignoring the different weights of the different regions will add uncertainty to the quality assessment of an image, since HVSs are different from the prediction processes generated by a CNN model.

Moreover, another two inevitable problems that arise in deep learning solutions are the shortage of data and the fixed-scale input required by an end-to-end model. Many pre-processing methods are deployed to solve these problems. However, these arbitrary ways will decrease the consistency between the images and their ground-truth scores, as shown in Figure~\ref{figure1}. The cropped images, like those in Figure~\ref{figure1}(b) and (f), show content clipped from original images. In detail, Figure~\ref{figure1}(b) includes the grass while the (a) contains both grass and an antelope. Also, the two cropped images from the same original image contain different levels and types of degradation, as shown in Figure~\ref{figure1}(f) and (g). Figure~\ref{figure1}(g) includes more severe blur than (f). Figure~\ref{figure1}(c) indicates that the rotation of one image will introduce unreal colorations and exerts a negative influence on the image quality. Rescaling deforms the subject of images thus affecting both the semantic content and quality of images, Figure~\ref{figure1}(d) and (h). As a result, the ground-truth quality is no longer suitable for the pre-processed images. Thus, using these pre-processed images to train a model will definitely cause bias in prediction and a less objective model.

In order to design a BIQA method that is more consistent with human perception, we propose a deep BIQA network  based on superpixels (DSN-IQA). Following the previous work~\cite{su2020blindly, lu2021blind}, we develope our CNN based network that extracts multi-scale semantic features, and fuse these features with the superpixel adjacency map obtained from superpixel segmentation model. Because when human eyes evaluate an image, they will pay attention to the semantic information and local details of the image at the same time and finally get the quality. Our network mimics this assessment process. In detail, we added superpixel segmentation to the method to help the network aware of local adjacency information. The images are segmented into superpixels, which are perceptually meaningful blocks, comprised of spatial neighboring pixels and generally given as a group of pixels. These pixels share similar local color and serve as low-dimensional representations of the images. Thus, they offer the detail information for further quality prediction. Consequently, our method simultaneously uses the local superpixels and multi-scale semantic features to ensure that the image evaluation process is more in line with the HVS. In addition, the method handles with the pre-processing problem by accepting arbitrary image sizes as input since humans assess the whole picture at the same time. 

We tested the model on multiple databases. In these experiments the test image remained at the original size so that the quality score of the test image accurately represents the truth quality. We conducted individual database, cross-database, individual distortion type, and ablation experiments. These test results show that the proposed adaptive model can handle complicated distortions and own high accuracy for predicting image quality due to the superpixel information extraction and sufficient multi-scale features.

In summary, our contributions are summarized as follows:
\begin{itemize}
	\item To the best of our knowledge, the proposed model is the first to apply superpixel segmentation in BIQA to extract local features and multi-scale semantic features. Experiments demonstrate that these features are highly consistent with HVS.
	
	\item We analyzed the influence of image cropping training on complete image evaluation, and adjusted the pooling layers to design a model that can overcome the problems associated with the size of images.
	
	\item Our approach is deployed properly and every part fits each other. The results of our experiments indicate that the approach outperforms in predicting quality and handles images with complicated distortions.
\end{itemize}

The remaining parts of this paper are arranged as follows. In Section~\ref{sect:related}, we review the development of superpixel segmentation for IQA and CNN-based BIQA models and show what our model builds from. Section~\ref{sect:Propose} details the construction of the proposed DSN-IQA model. Section~\ref{sect:Experi} provides extensive experimental results and a comparative analysis of our proposed model. Section~\ref{sect:conclu} summarizes our work and draws some conclusions.

\section{Related Works} \label{sect:related}
\subsection{Superpixel segmentation for IQA} \label{sect2:sup}
A superpixel, as defined by Ren \textit{et al.}~\cite{LCmodel} in 2003, refers to irregular pixel blocks with certain visual significance composed of adjacent pixels with similar texture, color, brightness and other characteristics. Superpixel segmentation uses a small number of visually meaningful superpixels to represent groups of adjacent similar pixels, reducing the volume of data. 

The widely used superpixel segmentation algorithms do not depend on CNNs relying instead on statistical models. These algorithms can be divided into two groups: the graph-based and the gradient-ascent-based algorithms. Graph-based algorithms are based on a data structure that contains vertices and weighted edges, and segments images by minimizing a cost function~\cite{felzenszwalb2004efficient, SLIC}. Some of the typical graph-based methods are Normalized cuts~\cite{NormalizedCuts}, Graph cuts~\cite{SPLattices}, and Entropy Rate Superpixel Segmentation algorithms~\cite{liu2011entropy}. Gradient-ascent-based algorithms are iterative and cluster pixels based on shifts between groups of pixels with similar values~\cite{SLIC}. There are a number of such segmentation algorithms currently in use, these include the Watershed~\cite{Watershedsindigital}, Mean Shift~\cite{MeanShift}, Quick Shift ~\cite{QuickShiftand}, Turbopixel~\cite{TurboPixels}, and Simple linear iterative clustering (SLIC)~\cite{SLIC} algorithms. SLIC in particular, achieves superpixel segmentation based on clustering pixels by color and distance similarities. Most of superpixel segmentation algorithms are unsupervised and produce superpixels with uniform size and regular shapes. These superpixels have clear visual meaning and are widely used in computer vision preprocessing.

IQA, includes two main improvements for superpixel assessment accuracy. First, superpixels reduce the pixel redundancy and help the automated assessment process be perceptive. Many common IQA algorithms use square convolution kernels to ensure that a sufficient number features are extracted from images for quality prediction ~\cite{kang2014convolutional, bianco2018use, kim2018deep}. The 3×3 square kernel however, concentrates on a tiny zone at a time and thus loses visual meaning~\cite{sun2018spsim}. As a result, a square kernel does not exploit the connections between adjacent pixels while promoting information redundancy. In contrast, superpixel segmentation performs much like human vision. When humans observe and assess a picture, adjacent similar pixels are recognized and gathered to one local region~\cite{shotton2009textonboost}. Second, superpixels help IQA models to assess regionally. Superpixel-Based Similarity Index (SPSIM)~\cite{sun2018spsim}, proposed by Sun \textit{et al.}, illustrates that different types of regions respond to noise distinctively. Textured areas perform more resistance to Gaussian noise than flat areas. The situation is reversed for image blur. If the extracting network ignores these meaningful effects in regions, it will locally lose some common details and the predicting result will definitely deviate from human’s subjective scores. All in all, superpixels are a vital tool for improving IQA algorithms.

Taking notice of superpixels, many superpixel-based methods emerged. SPSIM is a full-reference IQA model based on SLIC. In this method, reference images without distortion and distorted target images are segmented into visually meaningful regions, using superpixels. The mean values of the intensity and chrominance components are extracted within each superpixel and compared between a reference and target image to describe local similarity index precisely. Frackiewicz \textit{et al.} improved SPSIM and developed an improved SPSIM index for IQA~\cite{betterSPSIM}. Their method revises SPSIM in two ways: new color space replaces the YUV space and exploits the calculation method from Mean Deviation Similarity Index (MDSI) index~\cite{nafchi2016mean}. Fang \textit{et al.}~\cite{SuperpixelBasedQAoM} use SLIC to distinguish between the fused image and the exposed image. In this way, they compute the quality maps based on the Laplacian pyramid for large-changed and small-changed regions separately and take different regional strategies. All of these works bring regional solution to IQA by deploying superpixels. 

Superpixel implementation improves the consistency between algorithms and HVS, but there is a problem that still must be solved. It is difficult to combine non-CNN superpixel segmentation algorithms with CNN networks and ensure a better performance. CNN creates dimensional feature tensors which have their visual meanings. Thus, these tensors can’t combine simply with labeled superpixels created by non-CNN algorithms. At the same time, because CNN needs back propagation to train, these non-differentiable segmentation algorithms will block whole network and prevent training~\cite{jampani2018superpixel}. Nevertheless, directly using CNN to segment superpixel can overcome this contradiction. After comparing many other CNN superpixel segmentation methods, we opted for the Superpixel Segmentation Via CNN (SSVCNN) method proposed by Teppei~\cite{suzuki2020superpixel}. This is an unsupervised superpixel segmentation method that optimizes a randomly-initialized CNN. SSVCNN is easier to integrate with existing image quality models, and the output is a clear and meaningful probabilistic map representing the belonging of pixels.

\subsection{CNN-based BIQA models}
Due to the insufficient computational power, the early CNN-based IQA models cannot extract and predict quality in one network. These methods initially only extract hierarchical features and subsequent operations calculate quality from the feature sets~\cite{yang2020explicit}, which is different from an end-to-end model~\cite{ma2017end}. Tang \textit{et al.}~\cite{Tang_2014_CVPR} proposed a non end-to-end model using a radical basis function to pre-train a deep belief network with unlabeled data and fine-tune it with labeled data. In addition, Bianco \textit{et al.}~\cite{bianco2018use} adopted CNN features pre-trained on the image classification task as input in order to create quality evaluator through support vector regression (SVR)~\cite{drucker1997support}, a model that regresses quality scores from feature sets. These researchers quantified the mean opinion scores (MOS) into five categories, fine-tuned the pre-trained features in the multi category classification settings, and fed the fine-tuned features to SVR. This method however, cannot be optimized end-to-end since making many manual parameter adjustments necessary. 

Many end-to-end models started to emerge in BIQA however, with the development of computational power and deep CNN. Based on the BIQA model of CNN~\cite{Girshick_2014_CVPR}, Kang \textit{et al.} put forward the CNNIQA~\cite{kang2014convolutional} algorithm, which takes the image patch as input and employs back propagation and other methods for training. Since the feature extraction and regression are integrated into the CNN, the depth of the neural network is deepened to improve the learning ability. Kang \textit{et al.} proposed another algorithm CNNIQA++~\cite{Simultaneousefi}, which increased the number of convolutional layers of CNNIQA so that it simultaneously estimates image quality and distortion type. Zhang \textit{et al.} proposed the Deep Bilinear CNN (DBCNN) algorithm~\cite{zhang2018blind} based on VGG-16~\cite{simonyan2014very}. VGG-16 initially was designed for image recognition but can be fine-tuned to assess image quality. They designed two branches of deep CNNs, specializing in synthetically and authentically assessing distortion scenarios separately, using bilinear pooling to fuse the network of the two branches. These methods however, still cannot accurately predict authentic databases that contain many images with complicated objects.

In recent years, semantic features based BIQA models have become a research hotspot because of their ability to perceive semantic information and more accurately predict authentic image databases. Kim \textit{et al.}~\cite{kim2017deep} found ResNet ~\cite{he2016deep}, a deep semantic CNN trained by classification databases, also assists in improving accuracy in IQA. In~\cite{hosu2020koniq}, researchers tested several deep CNNs and confirmed the advantages of semantic features in dealing with authentic IQA databases. In Semantic Feature Aggregation method (SFA)~\cite{li2018has}, Li \textit{et al.} use statistics from ResNet-50 multi-patches features for quality prediction. They proposed that the content of images also has an impact on predicting the quality. They exemplified that people will score a clear blue-sky image as a high-quality one while the traditional prediction model mistakenly recognizes it as an image with blur noise. This phenomenon could be explained by the semantic losses during extracting features. Considering the contents varied from image to image, Su \textit{et al.} proposed Hyper-IQA~\cite{su2020blindly}. Hyper-IQA separates the IQA procedure into three stages including content understanding, perception rule learning and quality predicting. It designs a hyper network connection to mimic this mapping from image contents to the manner of perceiving quality. Thus, using semantic features in IQA will make an automated assessment process more like human assessment of the semantic contents in images when evaluating image quality.

The problem remaining is that all of the current semantic models lack attention to local visual meaning of images. As we illustrated in Section~\ref{sect2:sup}, superpixel fills this gap. Our proposed adaptive model takes a step further not only by using multi-scale semantic features, but also by introducing extraction of superpixel information to imitate HVS. These two innovations improve the consistency of assessment process between our model and human eyes.

\section{Proposed Method} \label{sect:Propose}
\subsection{Model framework} \label{sys framework}
Our method contains two models: multi-scale feature extraction and a superpixel network. In first model, we deploy a backbone network to extract multi-scale features including semantic information. In second model, we implement CNN based superpixel segmentation to make the deep neural network aware of local superpixel regions. In this way, we ensure that the fused features will contain the visual information from adjacent regions as provided by superpixels. By introducing these two models, our prediction process contains three steps:

\textbf{(1)} extract features with semantic information, 

\textbf{(2)} generate the superpixel adjacency map, 

\textbf{(3)} predict the quality score. 

\begin{figure*}[h!]
	\includegraphics[width = 0.95\textwidth]{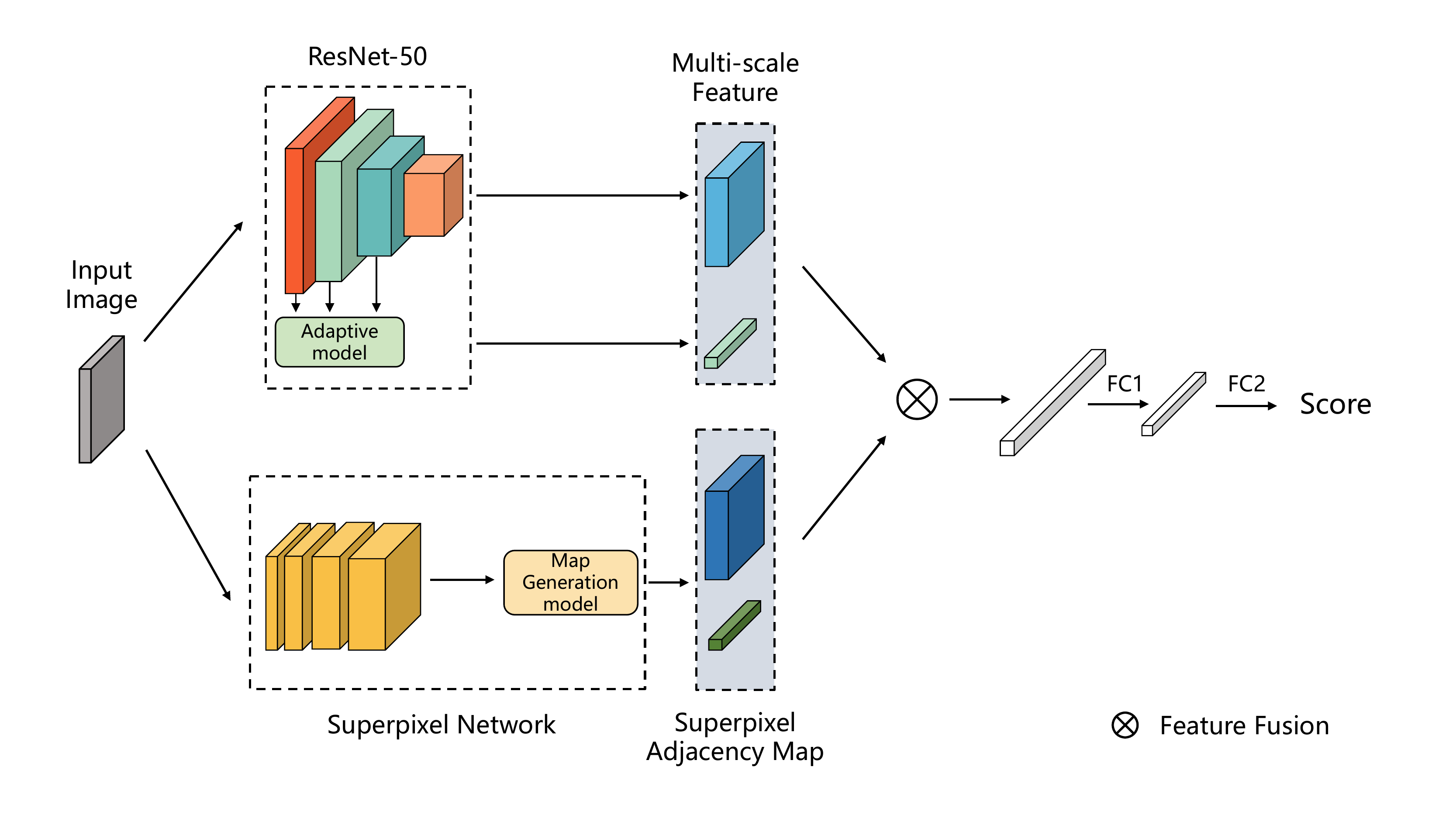}
	\includegraphics[width = .55\textwidth]{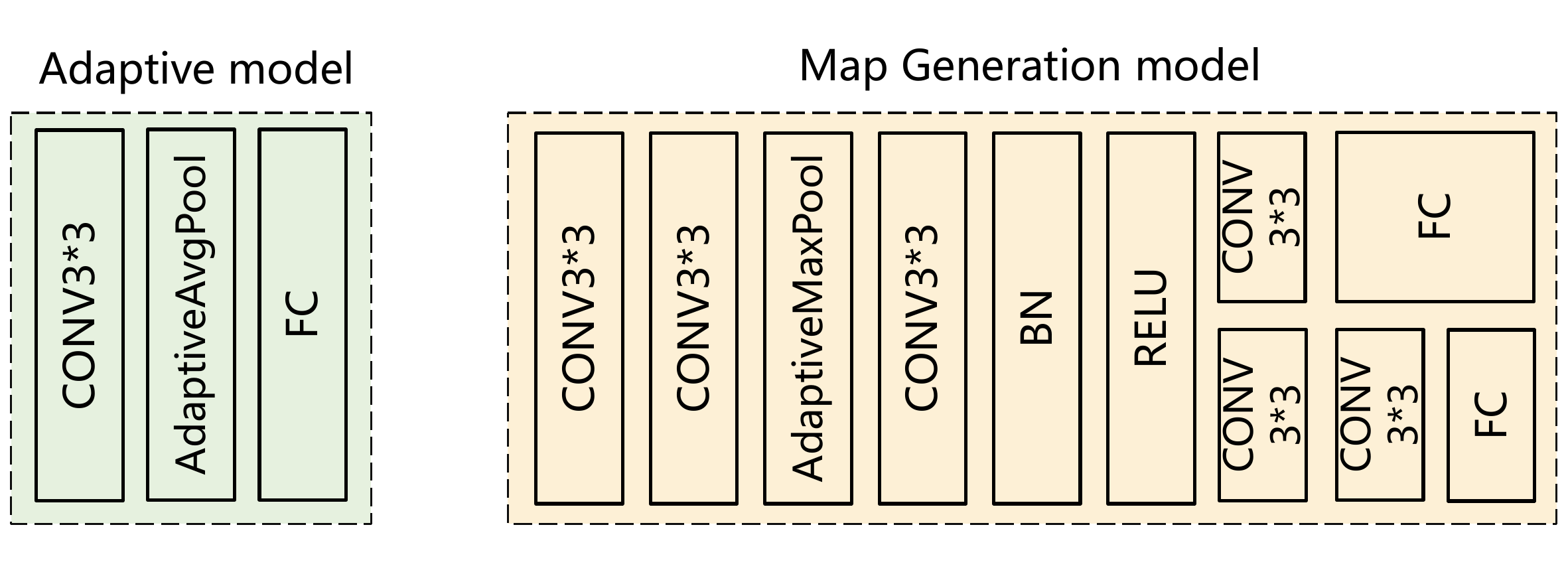}
	\caption{Structure of our proposed network.}
	\label{figure3}
\end{figure*}

The structure of the network is shown in Figure~\ref{figure3}. In detail, the input image is fed into one backbone network to extract semantic features and one CNN-based superpixel segmentation network to gain a superpixel probability map, respectively. For further consideration of features aggregation and refining the crucial part, we design a map generation network to gain a superpixel adjacency map after the segmentation. After generated, the superpixel adjacency map and the semantic features are combined together through features fusion. The mixed features are put into the prediction network to predict a final quality score. These mixed features, which are composed of semantic and local adjacent information, are highly comprehensive features and represent the image quality exactly. Consequently, our approach is consistent with the procedure of assessment, in which people concentrate on the semantic meaning and the local details of the image at the same time.

\subsection{Semantic Features Extraction} \label{FeatExtr}
We deploy pre-trained ResNet-50 as the backbone network to extract the semantic features and make the features more comprehensive. This network is aware of both the semantic contents and the quality of images. Inspired by earlier works~\cite{su2020blindly, lu2021blind},we apply a multi-scale features extraction model in our backbone network. In this way, the local contents and distortions are extracted more completely. Also, the multi-scale feature extraction strengthens the effect of superpixel adjacency map by conducting wider information fusion. Figure~\ref{figure3} shows the details of our multi-scale extraction model. We acquire the local features from the key points of the backbone network. In order to keep the principal component and fast speed of calculation, we apply a 1×1 convolution layer, an average pooling layer, and a full connection layer to refine the multi-scale features. With the introduction of multi-scale features, the network now can be defined as follows:

\begin{equation}
	V_{ms} = [L_1,L_2,L_3,F] = \varphi(\boldsymbol{x})
\end{equation}
where $ \boldsymbol{x} $ represents the input images, $ \varphi $ means the extraction model, $ L_{i}(i = 1,2,3) $ means the $ i $-th local feature and $ F $ means the holistic feature, and $ V_{ms} $ means the multi-scale features.

In addition, because our model is based on the consideration of semantic features and HVS, the input images should incorporate all of the original content. This requires that the pre-processing of input images should not modify the main contents and quality of image themselves. However, some researchers apply various augmented methods, including cropping the images to small patches unduly, resizing the images or padding the edges of images. All of these processes will change the content and quality of images, reducing the adjacent information thus making the MOS/DMOS labels unsuitable for the processed images. As a result, under the condition that the size varies from picture to picture, our model must be available for arbitrary size without affecting the quality. Although the convolution layers can accept arbitrary size images as input, FC layers only allow the fixed vectors, making the whole network accessible merely for fixed-size images. One common tactic is to use global average pooling (GAP)~\cite{Zhou_2016_CVPR, wu2021blind} or global maximum pooling (GMP)~\cite{su2020blind} to regularize the features. Although these approaches aim to establish the relation between the scalar amounts in features and channel amounts of the features, they lose too much information. For this reason, we displace the average pooling and maximum pooling with adaptive pooling, which GAP and GMP can be viewed as a special case of adaptive pooling. In this way, we adjust the size of features and save most of the useful information.

\subsection{Superpixel Map Generation Net} \label{SuperMap}
\subsubsection{Superpixel Segmentation} \label{SuperSeg}
SSVCNN defines the superpixel segmentation as N class classification problem. Network is devised by a five-layer CNN and the procedure of segmentation can be simply defined as:
\begin{equation}
	P = S(\boldsymbol{x})
\end{equation}
where $ P \in \mathbb{R}_{+}^{H \times W \times N}$; $ \sum_{n}P_{h,w,n}=1 $ is the probabilistic representation map of superpixels, $S(\boldsymbol{x})$ represents the whole network, $\boldsymbol{x}$ means the input image with size \textit{H}×\textit{W}. We delete the operation $ \arg\max_{n}P_{h,w,n} $, which is used to transform the superpixel probability map to visible superpixel for visual appreciation. In practical process, the visible superpixels are not necessary and thus we directly use the superpixel probability map. For implemented details, we mostly use the default parameters set by author. Particularly, we set the maximum number $N(n)$ of superpixel to 100 to accelerate the processing speed.

\begin{figure*}[h!]
	\centering
	\subfigure[Buildings]{
		\begin{minipage}[t]{\columnwidth}
			\centering
			\includegraphics[width=0.95\textwidth]{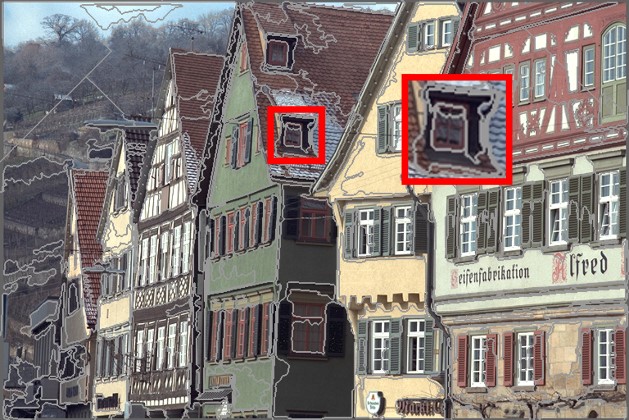}
		\end{minipage}%
	}%
	\subfigure[Church and Capitol]{
		\begin{minipage}[t]{\columnwidth}
			\centering
			\includegraphics[width=0.95\textwidth]{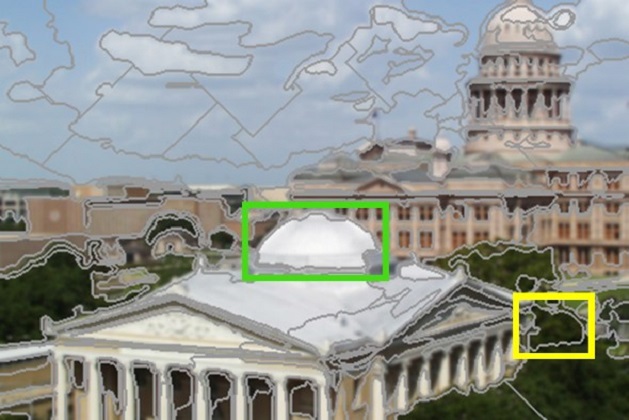}
		\end{minipage}%
	}%
	\caption{Exmaple of superpixel segmentation. (a) Buildings after superpixel segmentation. In red frame the window and the shadow of roof are separated, which demonstrated that superpixel always belongs to one semantic object. (b) Church and capito after superpixel segmentation. The gathered pixels of cupola share same smooth and flat characteristics. The clumped pixels of tree share similar texture. Both of them have different perception for human eyes collectively.}
	\label{figure2}
\end{figure*}

Figure~\ref{figure2} are two examples processed by our superpixel segmentation algorithm. The pictures in Figure~\ref{figure2} are divided into 100 superpixels and processed using the argmax function to present a crisper, clearer visual result. Adjacent pixels with the same type of features are grouped together in the image. These pixels are similar to each other in color and intensity and always belong to one semantic object. For instance, through the Figure~\ref{figure2}(a) we can clearly see that the area marked by red frame have semantic information implying a window. Also, the black shadow caused by roof and the body of window are separated clearly. Through Figure~\ref{figure2}(b) we can see that the area marked by green frame represents the roof and the area marked by yellow frame is part of a tree. The textured tree and the flat roof are successfully segmented and they have different resistance to blur noise~\cite{li2018has}. Thus, the information carried by those superpixels enables further IQA processing for enhanced visual results. This processing makes multi-scale semantic feature extraction conform more closely to quality of the picture for improved the adaptability and effectiveness of the algorithm. In this way, combining the result of superpixel segmentation with CNN can also make up for inability of existing algorithms to exploit the semantic content of images.

\subsubsection{Superpixel Map Generation} \label{SuperGer}
After the superpixel segmentation, we have the probabilistic representation map P with the size N×H×W. For future aggregation and reducing the redundancy, we designed a map generation model to generate a superpixel adjacency map. The structure of the network is shown in Figure~\ref{figure3}. We use several 3×3 convolution layers to gather meaningful features and then apply adaptive maximum pooling to make the network suitable for arbitrary sizes. According to the representation of features, two different branches are devised to fit the local features and holistic features, respectively. Owning the multi-scale features and superpixel adjacency map, we integrate them with direct multiplication and the mixed features are fed into FC layers to predict the final score.

\subsection{Model Training} \label{TrainPart}
\subsubsection{Implementation Detail} \label{Implement}
We deployed our model in PyTorch 1.7.1~\cite{paszke2019pytorch} and conducted training and testing in NVIDIA Tesla V100 GPUs 16G. For the consideration we stated in Section~\ref{FeatExtr}, we tested different size images according to their original sizes in database, but all size of training images is the same in one database. This is the requirement of mini-batch based training, which stabilizes the loss and increases the generalization ability. We used the Adam~\cite{kingma2014adam} optimizer with the weight decay rate $\lambda$ = 0.0005, contributing to avoiding over-fitting. It can be defined as following:

\begin{equation}
	\mathcal{L} = \mathcal{L}_{0}+\lambda \sum_{\omega}\omega^{2}
\end{equation}
where $ \mathcal{L} $  and  $ \mathcal{L}_{0} $ mean the patch loss and the original patch loss, and $ \omega $ means all training patches. Also, we set initial learning rate as $ 10^{-3} $  and applyed dynamic adjustment.

\subsubsection{Loss function} \label{Loss}
For convolution neural networks, stochastic gradient descent and backward propagation are widely used to calculate the gradient and update the parameters. Loss function acts as an index of the entire network. In our work, we minimize L1 loss, which describes the absolute error between predicted score and the ground-truth score over training set. It can be defined as follows:

\begin{equation}
	\ell=\frac{1}{M}\sum_{i}^{M}| \Phi(\varphi(\boldsymbol{x}_i),S(\boldsymbol{x}_i)) - {Q}_{i}|
\end{equation}
where $ \boldsymbol{x}_{i} $ and $ Q_i $ represent the $ i $-th training patch and its ground truth score, $ \Phi $ represents the prediction model, and $ M $ represents the number of input samples.

\section{Results and discussion} \label{sect:Experi}
In experiment stage, we selected 5 databases for training and testing. They are KonIQ-10K~\cite{hosu2020koniq}, LIVE In the Wild Image Quality Challenge (LIVEC)~\cite{ghadiyaram2015massive}, LIVE Facebook (FLIVE)~\cite{ying2020patches}, LIVE~\cite{sheikh2006statistical}, and CSIQ~\cite{larson2010most} database. All details are described below and shown in Table~\ref{t1}.

\begin{table*}[h]
	\centering
	\caption{Comparison of basic database conditions}
	\renewcommand{\arraystretch}{1.25}
	\label{t1}
	\scalebox{0.93}{
	\begin{tabular}{lllllll}
		\hline
		\textbf{Database}  & \textbf{Unique contents} & \textbf{Resolution}         & \textbf{Distortions} & \textbf{No. of distorted images} & \textbf{Distortion type} & \textbf{Quality Index} \\ \hline
		\textit{LIVEC}     & 1,162                     & 500 × 500 $\sim$ 640 × 960   & -                    & 1,162                             & in-the-wild              & MOS                    \\
		\textit{KonIQ-10k} & 10,073                      & 512 × 384 and 1,024 × 768               & -                    & 10,073                              & in-the-wild              & MOS                    \\
		\textit{FLIVE}     & 39,810                    & 160 ×  186 $\sim$ 1,200 × 660 & -                    & 39,810                            & in-the-wild              & MOS                    \\
		\textit{LIVE}      & 29                       & 480 × 720  $\sim$ 768 × 512  & 5                    & 779                              & synthetic                & DMOS                   \\
		\textit{CSIQ}      & 30                       & 512 × 512                   & 6                    & 866                              & synthetic                & DMOS                   \\ \hline
	\end{tabular}}
\end{table*}

The first three lines are the authentic distorted image databases, which aim to achieve the distribution of real distorted images. Compared to the rest two synthetic databases, their distortion types are more complex and composite, which always become a harder task for image quality prediction model. KonIQ-10K database is composed of 10, 073 distorted images in size 512×384 and 1024×768. They are chosen from a huge public database YFCC100m~\cite{thomee2016yfcc100m} and their MOS ranged in [0, 100] is provided as the ground truth. LIVEC database contains 1, 162 images in size 500×500 created by cameras in the real world. MOS ranged in [0,100] is used and collected through online platform. The LIVE Facebook database containing 39, 810 images selected from several databases, was released and implemented on the Amazon Mechanical Truck (AMT) crowd sourcing system to gather each image MOS, ranged in the [0, 100].

In addition, we also used synthetic databases to test our work. The LIVE database has 779 synthetically distorted images and provides DMOS. CSIQ database provides 866 images and takes DMOS ranged in [0,1] as ground truth score 

Two commonly used criteria, namely Spearman’s rank order correlation coefficient (SRCC) and Pearson’s linear correlation coefficient (PLCC) were selected to evaluate the model. SRCC represents the monotonicity of the algorithm and PLCC represents the accuracy of prediction. Both range from -1 to 1 and a higher value indicates a model more consistent with human eyes.

We conducted the following splitting approaches in several experiments, including individual database, individual distortion type, ablation, and sub-image size experiment. From the authentic distorted image databases, KonIQ-10k, and LIVEC we randomly selected 80\% images as training set and 20\% images as testing set. In synthetic image databases, we also used the 8:2 train-test ratio to split the reference images, contributing to making the contents of image independent in the training and testing set.

\subsection{Performance on individual database} \label{Indi}
The proposed model was trained and tested on the same database split by method detailed in Section~\ref{sect:Experi}. The splitting and train-test procedures were repeated 10 times and the median results are provided. This decreases chance results and avoid dependence of the model on the specific training set. We compared our model with 10 other approaches, including:
\begin{itemize}
	\item Full reference based approaches: PSNR, SSIM~\cite{wang2004image}. Hand-crafted features based BIQA approaches: BRISQUE~\cite{mittal2012no}, BMPRI~\cite{min2018blind}.
	\item  Deep learning based BIQA approaches: CNN-IQA~\cite{kang2014convolutional}, WaDIQaM-NR~\cite{bosse2017deep}, SFA~\cite{li2018has}, DIQA~\cite{kim2018deep}, HFANet~\cite{wu2021blind}, DBCNN~\cite{zhang2018blind}.
\end{itemize}
Particularly, for some deep learning-based methods, it is difficult to reproduce the result due to the unavailability of code and related parameters so their results were extracted from the corresponding papers~\cite{zhang2018blind,  wu2021blind, sun2021blind, 9271914} for comparative analysis. The comparative results are shown in Table~\ref{t2} and the best performances in each database are shown in bold.

\begin{table}[h!]
	\centering
	\caption{Overall Performance on individual database}				
	\renewcommand{\arraystretch}{1.25}		
	\label{t2}		
	\scalebox{0.92}{						
		\begin{tabular}{l|lllll}
			\hline
			SRCC                & \textbf{LIVEC} & \textbf{KonIQ10k} & \textbf{FLIVE} & \textbf{LIVE}  & \textbf{CSIQ}  \\ \hline
			\textit{PSNR}       & -              & -                 & -              & 0.876          & 0.806          \\
			\textit{SSIM}       & -              & -                 & -              & 0.913          & 0.834          \\
			\textit{BRISQUE}    & 0.601          & 0.715             & 0.320          & 0.942          & 0.698          \\
			\textit{BMPRI}      & 0.487          & 0.658             & 0.274          & 0.931          & 0.908          \\
			\textit{CNN-IQA}    & 0.627          & 0.685             & 0.306          & 0.955          & 0.683          \\
			\textit{WaDIQaM-NR} & 0.692          & 0.710             & 0.452          & 0.954          & -     \\
			\textit{SFA}        & 0.804          & 0.888             & 0.542          & 0.883          & 0.796          \\
			\textit{DIQA}       & 0.703          & -                 & -              & \textbf{0.970}          & 0.844          \\
			\textit{HFANet}     & 0.754          & -                 & -              & 0.950          & 0.913          \\
			\textit{DBCNN}      & 0.851          & 0.875             & \textbf{0.554}          & 0.968          & \textbf{0.946}          \\
			\textbf{DSN-IQA}    & \textbf{0.854} & \textbf{0.913}    & 0.526 		  & 0.954	       & 0.921 \\ \hline
			
			PLCC                & \textbf{LIVEC} & \textbf{KonIQ10k} & \textbf{FLIVE} & \textbf{LIVE}  & \textbf{CSIQ}  \\ \hline
			\textit{PSNR}       & -              & -                 & -              & 0.872          & 0.800          \\
			\textit{SSIM}       & -              & -                 & -              & 0.945          & 0.861          \\
			\textit{BRISQUE}    & 0.621          & 0.702             & 0.356          & 0.935          & 0.829          \\
			\textit{BMPRI}      & 0.523          & 0.655             & 0.315          & 0.933          & 0.934          \\
			\textit{CNN-IQA}    & 0.601          & 0.684             & 0.285          & 0.953          & 0.754          \\
			\textit{WaDIQaM-NR} & 0.730          & 0.738             & 0.433          & 0.963          & -              \\
			\textit{SFA}        & 0.821          & 0.897             & 0.626          & 0.895          & 0.818          \\
			\textit{DIQA}       & 0.704          & -                 & -              & \textbf{0.972}          & 0.880          \\
			\textit{HFANet}     & 0.766          & -                 & -              & 0.963          & 0.918          \\
			\textit{DBCNN}      & 0.869          & 0.884             & \textbf{0.652}          & 0.971          & \textbf{0.959}          \\
			\textbf{DSN-IQA}   & \textbf{0.880} & \textbf{0.932}    & 0.623 & 0.954 & 0.910 \\ \hline
	\end{tabular} }

\end{table}

On authentic image databases, our model outperformed the other tested algorithms. These results meet our expectation that our visual system-based model can handle the complex situation in authentic images. Moreover, compared to synthetic databases, authentic databases contain more different semantic contents and thus have less repetition. This suggests that with more data driving, the model is effective and avoids over-fitting. Among all of the other approaches, Deep Bilinear CNN (DBCNN) also outperformed in FLIVE database. Because it contains two branches for dealing with authentic and synthetic images. One of the branches is pre-trained by PASCAL VOC 2012 database~\cite{everingham2010pascal} which is also used by FLIVE database partially. As a result, it brings some advantages to DBCNN when it trains and tests in FLIVE.

For synthetic databases, although our model is not intended for them originally, it still gained the top three SRCC values on CSIQ database and above the average on LIVE. This illustrates that our model gives the high score to high quality images and gives low score to bad quality images correctly. But the predicted scores are not as accurate as the ground-truth scores in label.

\subsection{Performance on cross-database} \label{Cross}
Cross-database experiment was designed to test the generalization ability of our approach. Cross-database experiment trains the model on one database and tests it on another independent database. A robust model can not only perform effectively on training database, but also on other databases. The whole cross-database test is separated into two parts, an authentic part and a synthetic part. We should mention that because of the distorted types, processes of taking the image and the distribution of ground truth are different in each database, the evaluation metrics will certainly degrade, especially when the type of database is different.

In the authentic distorted databases, we selected the most competing approach DBCNN to compare. We test the models with other two authentic databases, which can bring a sufficient view of the generalization ability. For FLIVE database, we follow the process in~\cite{ying2020patches} to collect the testing set. It is excluded from the training set because of its specific requirement to the pre-processing. Table~\ref{t3} shows the SRCC result in authentic part.
\begin{table}[h]
	\centering
	\renewcommand{\arraystretch}{1.25}
	\setlength{\abovecaptionskip}{10pt}%
	\caption{Result of SRCC compared with DBCNN on authentic cross-database}
	\label{t3}
	\scalebox{1.0}{
		\begin{tabular}{c|ccc}
			\hline
			\diagbox {Train}{Test} 			& KonIQ-10k            & LIVEC                & FLIVE\_test          \\ \hline
			KonIQ-10k       & \textbf{0.913}/0.875 & \textbf{0.780}/0.755  & \textbf{0.470}/0.470   \\ 
			LIVEC           & 0.724/\textbf{0.754}          & \textbf{0.854}/0.851 & \textbf{0.448}/0.405 \\ 
			\hline
	\end{tabular}}
\end{table}
The left values are provided by our method and the right values are calculated by DBCNN. The better method’s index between them is highlighted in bold text. Our model outperformed 3 out of 4 experiments, which illustrates that our trained model is applicable for more different images. For our model trained by two authentic databases, comparing the test result in FLIVE suggests that the model assesses more perceptively and widely if there are more data for training. Thus, we can ensure the sufficient images for training to improve the generalization ability. 

One synthetic database, became a training set and the other was used for testing. We also added LIVEC as the extended testing set.  For clearer comparison, we pick 6 outstanding approaches, including BLIINDS-II~\cite{saad2012blind}, CNN-IQA, BIECON~\cite{kim2016fully}, PQR~\cite{zeng2017probabilistic}, WaDIQaM-NR, and TTL-IQA~\cite{9271914}.The SRCC results are shown in Table~\ref{t4}. The top two indicators of generalization ability are shown in bold.

\begin{table*}[h!]
	\centering
	\renewcommand{\arraystretch}{1.25}
	\caption{SRCC Results on synthetic cross-database}
	\label{t4}
	\begin{tabular}{c|c|ccccccc}
		\hline
		Train& Test & \textbf{DSN-IQA} & \textit{BLIINDS-II} & \textit{CNN-IQA} & \textit{BIECON} & \textit{PQR}   & \textit{WaDIQaM-NR} & \textit{TTL-IQA} \\ \hline
		LIVEC & LIVE  & \textbf{0.603}    & 0.228      & 0.427  & 0.435  & 0.440  & -          & \textbf{0.676}   \\	
		& CSIQ  & \textbf{0.623}    & 0.305      & 0.239  & 0.412  & 0.538 & -          & \textbf{0.606}   \\	\hline
		LIVE  & LIVEC & \textbf{0.520}     & 0.369      & 0.400    & 0.427  & \textbf{0.547} & -          & 0.490    \\	
		& CSIQ  & \textbf{0.742}    & 0.601      & 0.723  & 0.719  & 0.717 & 0.704      &\textbf{ 0.807}   \\ \hline
		CSIQ  & LIVE  & 0.896    & 0.894      & 0.854  & \textbf{0.922}  & \textbf{0.930}  & -          & 0.840    \\ 
		& LIVEC & \textbf{0.436}    & 0.276      & 0.300    & 0.317  & \textbf{0.479} & -          & 0.298  \\ 
		\hline
	\end{tabular}

\end{table*}

Our model achieved the top performance of generalization ability. When we focus on the are comparatively harder experiments which contain LIVEC database as training set or test set, our model stands out every time. This verifies that our model owns high generalization ability since the type of LIVEC database is different from two synthetic databases. Also, the results reflect that the models trained and tested in same type databases gain more precise performance. As the data distributions are similar, the models can get considerable ability for generalization.

\subsection{Performance on each distortion type} \label{Each Type}
After testing on holistic database, we analyze the results of each distortion type separately. This experiment measures the ability for assessing the quality of specific distortion. We only conducted the experiments on the synthetic databases because the distortion types are extremely complex in authentic databases and also not tagged. The models were trained in database with all types of distorted images and tested by specific distorted type. Table~\ref{t5} shows the result from LIVE and CSIQ database and the top three performance values are in bold. From these two tables, we can observe that our model is among the top two performing models, showing a significant advantage in dealing with usual distortions. Especially, our model outperforms in Gaussian Blur and Fast Fading. Due to the superpixel segmentation, we can extract their adjacency information more adequately although their contents have suffered from a severe distortion. Thus, our model has ability to gain semantic features and evaluate more accurately for these two distortions. However, our model did not perform as well as others on JPEG and JP2K images. Both contaminate the surrounding pixels severely thus have negative impacts on superpixel segmentation. Under this situation, thanks to the semantic and multi-scales features, our approach still has appreciable results.

\begin{table*}[h!]
	\centering
	\renewcommand{\arraystretch}{1.25}
	\setlength{\abovecaptionskip}{10pt}%
	\caption{SRCC Results of specific distortion types on LIVE database and CSIQ database}
	\label{t5}
	\begin{tabular}{l|llllll|lllllll}
		\hline
		Database            & \multicolumn{6}{c}{LIVE}                                                                            & \multicolumn{7}{c}{CSIQ}                                                                                             \\ \hline
		Type           & \textbf{JP2K}  & \textbf{JPEG}  & \textbf{WN}    & \textbf{GB}    & \textbf{FF}    & \textbf{ALL}   & \textbf{JP2K}  & \textbf{JPEG}  & \textbf{WN}    & \textbf{GB}    & \textbf{PN}    & \textbf{CC}    & \textbf{ALL}   \\ \hline
		\textit{BRISQUE}    & 0.929          & 0.965          & \textbf{0.982} & \textbf{0.964} & 0.828          & 0.939          & 0.840           & 0.806          & 0.723          & 0.820           & 0.378          & 0.804          & 0.746          \\ 
		\textit{BLIINDS-II} & 0.930           & 0.950           & 0.947          & 0.915          & 0.871          & 0.912          & 0.850           & 0.846          & 0.702          & 0.880           & 0.812          & 0.336          & 0.780           \\
		\textit{BIECON}     & 0.952          & \textbf{0.974} & 0.980           & 0.956          & 0.923          & 0.961          & \textbf{0.954} & \textbf{0.942} & 0.902          & \textbf{0.946} & 0.884          & 0.523          & 0.815          \\
		\textit{PQR}        & 0.953          & 0.965          & 0.981          & 0.944          & 0.921          & \textbf{0.965} & \textbf{0.955} & \textbf{0.934} & 0.915          & 0.921          & 0.926          & 0.837          & 0.873          \\
		\textit{BMPRI}      & 0.939          & 0.967          & \textbf{0.986} & 0.918          & 0.827          & 0.931          & 0.900            & 0.918          & \textbf{0.928} & 0.918          & -              & -              & 0.909          \\
		\textit{DIQA}       & \textbf{0.961} & \textbf{0.976} & \textbf{0.988} & \textbf{0.962} & 0.912          & \textbf{0.975} & 0.927          & 0.931          & 0.835          & 0.870           & 0.893          & 0.718          & 0.884          \\
		\textit{DBCNN}      & \textbf{0.955} & \textbf{0.972} & 0.980           & 0.935          & \textbf{0.930}  & \textbf{0.968} & 0.953          & \textbf{0.940}  & \textbf{0.948} & \textbf{0.947} & \textbf{0.940}  & \textbf{0.870}  & \textbf{0.946} \\
		\textit{HyperIQA}   & 0.949          & 0.961          & \textbf{0.982} & 0.926          & \textbf{0.934} & 0.962          & \textbf{0.960}  & 0.934          & \textbf{0.927} & 0.915          & \textbf{0.931} & \textbf{0.874} & \textbf{0.923} \\
		\textbf{DSN-IQA}    & \textbf{0.956} & 0.933          & 0.975          & \textbf{0.967} & \textbf{0.955} & 0.954          & 0.936          & 0.931          & 0.886          & \textbf{0.944} & \textbf{0.929} & \textbf{0.869} & \textbf{0.921} \\ \hline
	\end{tabular}
\end{table*}

\subsection{Ablation experiment} \label{AblExp}
In order to verify the improvement of our approach, we set 4 ablation experiments. First, we select the ResNet50 as basic experiment to evaluate the ability of baseline. Resnet50-ft accepts fixed size 224×224 as input and is fine-tuned to be tested on same size images. For the Resnet50-arbitrary, the simple adaptive model is added and thus tested on arbitrary size images. Secondly, we only deploy the multi-scale extraction model to the backbone. The result can suggest whether it can improve the ability of baseline network. At last, our model is tested to demonstrate that a combination of two sub models is appropriate for the whole network. All SRCC and PLCC experiments are shown in Table~\ref{t7}. The best results are emphasized by bold and the letters behind the database name indicate the SRCC (S) and PLCC (P) metric.
\begin{table}[h!]
	\centering
	\renewcommand{\arraystretch}{1.25}
	\caption{Ablation experiment}
	\label{t7}
	\scalebox{0.90}{
		\begin{tabular}{l|llll}
			\hline
			\textbf{Method}    & \textbf{LIVEC\_S} & \textbf{LIVEC\_P} & \textbf{LIVE\_S} & \textbf{LIVE\_P} \\ \hline
			Resnet50-ft   & 0.819             & 0.849             & 0.954            & 0.950            \\ 
			Resnet50-arbitrary & 0.848             & 0.869             & 0.946            & 0.939            \\ 
			Resnet50+Multi     & 0.850             & 0.872             & 0.952            & 0.946            \\ 
			Resnet50+Multi+SP  & \textbf{0.855}    & \textbf{0.880}    & \textbf{0.954}   & \textbf{0.954}   \\ \hline
	\end{tabular}}
\end{table}

Table~\ref{t7} indicates the function of each specific model. The first-row result, taken from~\cite{8103112}, shows that the mere usage of a semantic network can outperform many hand-crafted and CNN based approaches. Compared to the first row, the method on the second row accepts the intact images and thus saves more semantic information for the assessment process. The result on second method shows 2-3\% improvement on LIVEC after only allowing arbitrary image sizes. Due to the special component of image size in LIVE and the requirement of same image size in one mini-batch, the arbitrary model cannot perform fully in LIVE. The results found in row 3 and 4, definitely indicate that both of the sub models fit each other and help the network achieve better performance. Multi-scale features extraction, with its awareness of local and global features, and fits human assessment precisely. The superpixel segment model permits the extraction of regional contents and simulates the human visual system. The result in this part illustrates that the consolidation of multi-scale and superpixel segmentation is a feasible and effective way to extract the accurate features of image quality.

\subsection{Effect of sub-image size} \label{PatSize}
Considering the limited image amounts and various image sizes in some databases, sometimes cropping the image to specific size for data augmented and training is necessary. However, the quality distribution within an image is regional and uneven. Every random cropping for the same image will create sub-images that vary in quality. As a result, there must be a sub-image size that minimizes distortion. 

\begin{figure}[h]
	\includegraphics[width = .45\textwidth]{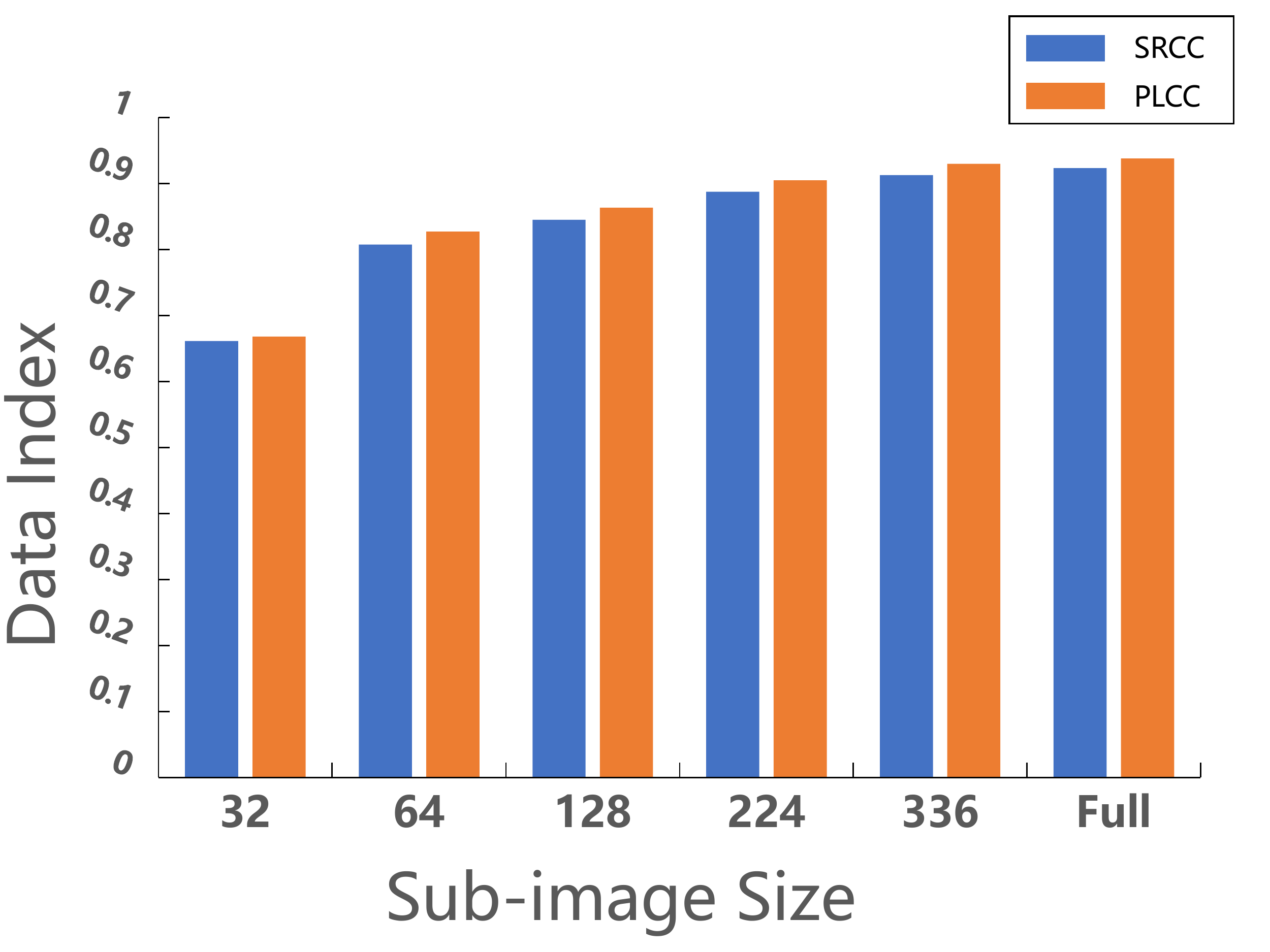}
	\caption{Comparison histogram of sub-image size results in LIVEC}
	\label{histogram1}
	\includegraphics[width = .45\textwidth]{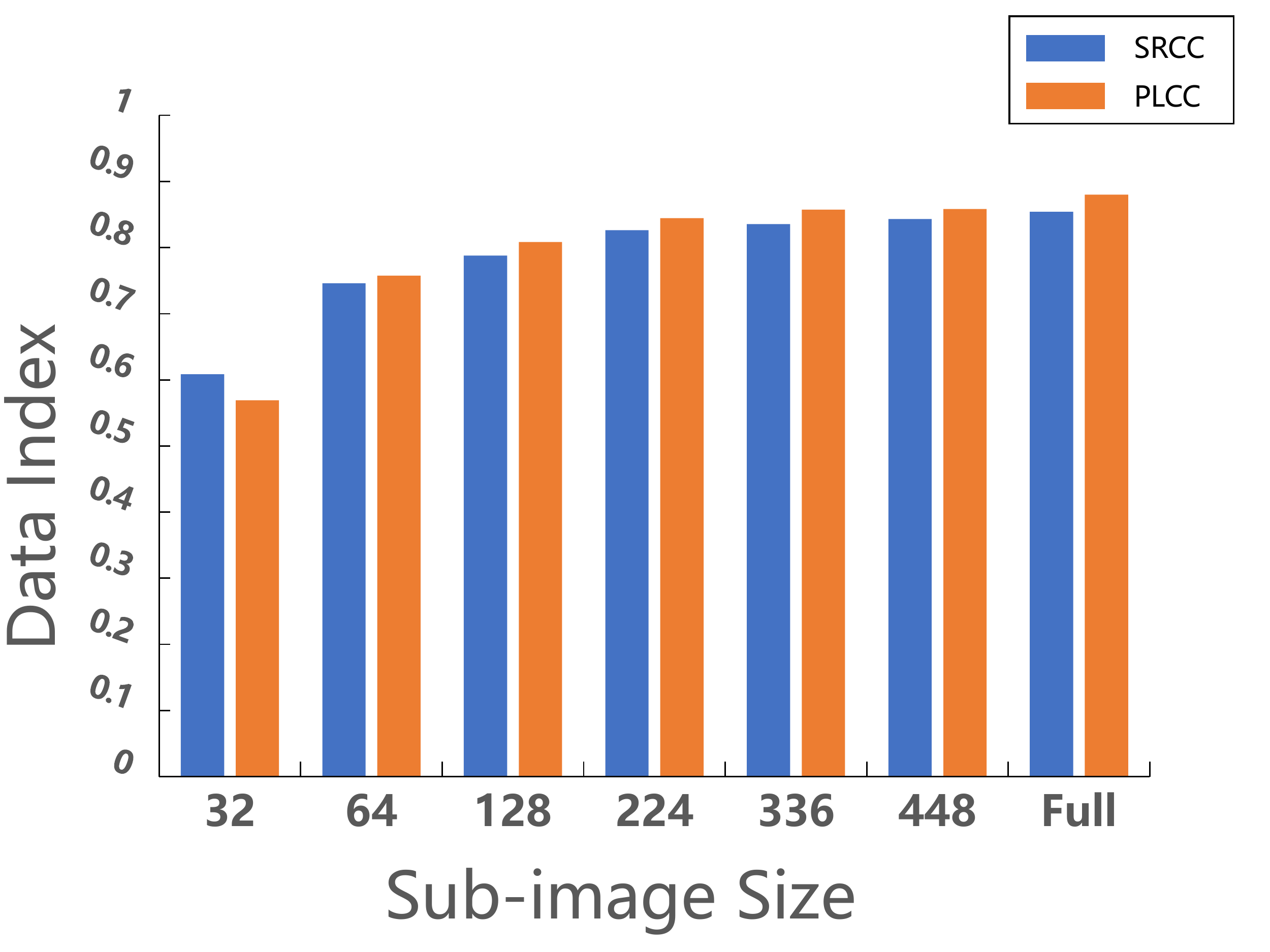}
	\caption{Comparison histogram of sub-image size results in KonIQ-10k}
	\label{histogram2}
\end{figure}

We designed the following experiment to determine the optimal sub-image size. For this experiment we chose the LIVEC and KonIQ-10k databases because all image sizes are same. Thus, we can control the variate. For training set, we randomly cropped~\cite{paszke2019pytorch} the images into various sizes, ranging from 32×32 to full size. The total epoch was adjusted by the size to ensure the sufficient training. For the testing set, all the experiments were evaluated using original size images. The results are shown in Figure~\ref{histogram1} displaying the results from LIVEC and Figure~\ref{histogram2}, depicting those from the KonIQ-10k database. 
It shows that when the size grows larger, the monotonicity and accuracy of models simultaneously increased. This phenomenon demonstrates that the model can perceive the quality correctly and evaluate arbitrary size images precisely with more training contents being saved. Thus, ensuring the consistency and conformity between training set and testing set is necessary for the network to be fully trained and deal with complicated images.

\section{Conclusion} \label{sect:conclu}
In this paper, we propose an BIQA model based on multi-scale semantic features and superpixels. For the input images, our improved pooling approach avoids changes in the quality of input images caused by pre-processing methods and thus arbitrary scale images can be accepted by the proposed model, and the original information and quality are preserved. In our proposed model, multi-scale features containing semantic and quality information are gathered by the backbone network. These multi-scale features mimic the information produced by human eyes when people assess one image and thus can predict a credible result. Furthermore, since the adjacent pixels share many similar attributes and have a certain impact on perception, we implement a superpixel model to extract the neighboring information. This information also contains semantic information and complements multi-scale features. As a result, with the fusion of these two elements, the prediction model is highly consistent with human perception and solves complicated images.

The proposed model tackles complicated authentic images and accepts arbitrary size images as input during testing period. How to make the training period accept any size images and explore a more efficient way to exploit superpixels could be the starting point for further research.


\begin{backmatter}
	
\section*{Abbreviations}
AMT: Amazon Mechanical Truck;
BIQA: Blind image quality assessment;
CC: Contrast noise;
CNN: Convolutional neural network;
CSIQ: Categorical subjective image quality;
DMOS: Differential mean opinion score;
DSN: Deep superpixel-based network;
FC: Full connected;
FF: Fast fading;
FLIVE: LIVE facebook;
FR-IQA: Full-reference IQA;
ft: fine-tune;
GAP: Global average pooling;
GB: Gaussian blur;
GMP: Global maximum pooling;
HVS: Human visual system;
IQA: Image quality assessment;
JP2K: JPEG 2000;
JPEG: Joint Photographic Experts Group;
LIVE: Laboratory for image and video engineering;
LIVEC: LIVE in the wild image quality challenge;
MOS: Mean opinion scores;
MS: Multi-scale;
NR-IQA: No-reference IQA;
NSS: Natural scene statistics;
PLCC: Pearson's linear correlation coefficient;
PN: Pink noise;
ResNet: Residual network;
RR-IQA: Reduced-reference IQA;
SFA: Semantic Feature Aggregation;
SLIC: Simple linear iterative clustering;
SPSIM: Superpixel-Based Similarity Index;
SRCC: Spearman's rank order correlation coefficient;
SSVCNN: Superpixel Segmentation Via CNN;
SVR: Support vector regression;
WN: White noise;

\section*{Acknowledgements}
The numerical calculations in this paper have been done on the supercomputing system in the Supercomputing Center of Wuhan University.

\section*{Authors' contributions}
ZY implemented the core method, performed the statistical analysis and drafted the manuscript. YX conducted the experiments and drafted the manuscript. GY designed the methodology. 
All authors read and approved the final manuscript. 

\section*{Funding}
This study is partially supported by National Natural Science Foundation of China (NSFC) (No. $61871298$, $42071322$) and National Key Research and Development Program of China (No. $2018YFB0504501$).

\section*{Authors' information}
The authors are with the School of Electronic Information, Wuhan University, Wuhan 430072, China.

\section*{Availability of data and materials}
The Python source code of DSN-IQA can be downloaded
at~\href{https://github.com/SN-F-QR/DSN-IQA}{https://github.com/SN-F-QR/DSN-IQA} for public use and
evaluation. You can change this program as you like and use it anywhere, but please refer to its original source.

\section*{Competing interests}
The authors declare that they have no competing interests.

\bibliographystyle{bmc-mathphys} 
\bibliography{bmc_article}      







\end{backmatter}
\end{document}